\documentclass{article}
\usepackage{ijcai24}

\usepackage{times}
\usepackage{soul}
\usepackage{url}
\usepackage[hidelinks]{hyperref}
\usepackage[utf8]{inputenc}
\usepackage{caption}
\usepackage{graphicx}
\usepackage{amsmath}
\usepackage{amsthm}
\usepackage{booktabs}
\usepackage[switch]{lineno}
\usepackage{xcolor}

\usepackage{amssymb}

\usepackage[utf8]{inputenc}
\usepackage{amsmath,amsfonts}
\usepackage{graphicx}
\usepackage{textcomp}

\usepackage{multirow}
\usepackage{times}
\usepackage{soul}
\usepackage{url}
\usepackage[hidelinks]{hyperref}
\usepackage[utf8]{inputenc}
\usepackage{caption}
\usepackage{algpseudocode}
\usepackage{wrapfig}
\usepackage{lscape}
\usepackage{rotating}
\usepackage{subcaption}

\title{Deep Neural Networks via Complex Network Theory: a Perspective}

\author{
\footnote{Co-first authors. \\ \indent Correspondence at emanuele.lamalfa@cs.ox.ac.uk.}Emanuele La Malfa$^1$
\and
$^*$Gabriele La Malfa$^2$\and
Giuseppe Nicosia$^{3}$\And
Vito Latora$^{3,4}$\\
\affiliations
\small{$^1$University of Oxford\\
$^2$King's College London\\
$^3$University of Catania\\
$^4$Queen Mary University of London}\\
\emails
\small{\{emanuele.lamalfa@cs.ox.ac.uk \
gabriele.la\_malfa@kcl.ac.uk \ 
giuseppe.nicosia@unict.it \
v.latora@qmul.ac.uk\}
}
}

\begin{document}

\maketitle

\begin{abstract}
Deep Neural Networks (DNNs) can be represented as graphs whose links and vertices iteratively process data and solve tasks sub-optimally. Complex Network Theory (CNT), merging statistical physics with graph theory, provides a method for interpreting neural networks by analysing their weights and neuron structures. However, classic works adapt CNT metrics that only permit a topological analysis as they do not account for the effect of the input data. In addition, CNT metrics have been applied to a limited range of architectures, mainly including Fully Connected neural networks. 
In this work, we extend the existing CNT metrics with measures that sample from the DNNs' training distribution, shifting from a purely topological analysis to one that connects with the interpretability of deep learning. For the novel metrics, in addition to the existing ones, we provide a mathematical formalisation for Fully Connected, AutoEncoder, Convolutional and Recurrent neural networks, 
of which we vary the activation functions and the number of hidden layers. We show that these metrics differentiate DNNs based on the architecture, the number of hidden layers, and the activation function. Our contribution provides a method rooted in physics for interpreting DNNs that offers insights beyond the traditional input-output relationship and the CNT topological analysis. 
\end{abstract}

\section{Introduction}~\label{S1}
Deep Neural Networks (DNNs) are learning algorithms loosely inspired by the human brain: they consist of layers of interconnected nodes called \emph{neurons} that process an input to produce an output~\cite{BISHOP1995,SCHMIDHUBER2015,GOODFELLOW2016}.
Such models perform remarkably on many tasks without requiring humans to engineer features manually, as each DNN layer delegates neurons to learning and representing specific features, the so-called \emph{hierarchical representation} of the input~\cite{singh2018hierarchical}. However, the outstanding ability of DNNs to learn these representations is accompanied by the challenge of \emph{interpreting} both what happens inside the neural networks and the mapping process between a representation (data) and its label (output)~\cite{montavon2018methods,ghorbani2019interpretation}. 
Since the onset of the modern era of machine learning, interpreting the learning mechanisms of neural networks has emerged as a primary area of research, as highlighted in seminal works such as Erhan et al., Karpathy, and Zeiler et al.~\shortcite{ERHAN2009,KARPATHY2015,ZEILER2014}. In this work, we seek to unify, understand and represent DNNs and their dynamics through the lens of graph models. Graph models works include, among the others, Graph-Based Models~\cite{Kipf2017}, Layer-wise Relevance Propagation~\cite{Montavon2019}, and Complex Network Theory~\cite{BOCCALETTI2006}. In particular, Complex Network Theory (CNT) is a branch of mathematics that represents complex systems, from city connectivity to networks of computers~\cite{porta2006network}, by modelling and then simulating their dynamics through graphs where nodes represent entities and vertices relationships~\cite{CRUCITTI2004,PORTA2006,CHAVEZ2010}. 
CNT offers an intuitive method for conceptualising DNNs, where neurons are analogous to graph nodes and connections to weighted edges. 

This paper characterises DNNs as graphs via CNT metrics: we develop and unify a set of metrics that describe a network's weights, neurons, and the hidden layers' behaviour: we uncover consistent trends across various architectures, initialisations, and objective tasks across range of network architectures, including Fully Connected (FC), AutoEncoders (AE), Convolutional (CNNs), and Recurrent Neural Networks (RNN). 
The visual representations of the CNT metrics provide insights into a DNN's decision process.
In synthesis, our work formally connects and grounds the existing work in the field~\cite{TESTOLIN2019,SCABINI2022} by unifying and extending existing CNT metrics such as \emph{Link Weigths Dynamics}, \emph{Nodes Strength} and \emph{Layers Fluctuation}~\cite{LAMALFA2021} moving beyond a topological analysis to account for the effect of the input data.
Section~\hyperref[S2]{2} surveys the literature of CNT applied to DNNs. In Section~\hyperref[S3]{3}, we introduce our methodology, i.e., a unified framework to study DNNs via CNT metrics and how to compute the various metrics for different DNN architectures. Section~\hyperref[S4]{4} reports results for different architectures, activation functions, and shallow and deep networks. We conclude the article with some ideas for further development and future works in this raising area of research.

\section{Related Works}~\label{S2}
Testolin et al.~\shortcite{TESTOLIN2019} first used CNT to interpret deep learning models on classification tasks. Their work leverages CNT to retrieve information from Deep Belief Networks, a generative model whose unsupervised learning phase differs from feed-forward neural networks. They conducted experiments on MNIST and discovered a tension between elicited and suppressed neurons when classifying digits and that suppression is correlated with the efficiency of a network on the classification task.
Zambra et al.~\shortcite{ZAMBRA2020} moved beyond neuron activations and studied DNNs connected components, or `motifs' as connectivity patterns between neurons or layers. `Motifs' shed light on how a DNN learns and generalises. They discovered that initialising a neural network weights is key to the emergence of such `motifs' and connects to accurate learning.
In their study, Petri et al.~\shortcite{PETRI2021} show the inherent trade-offs between the capacity of a DNN to learn, generalise and execute multiple tasks simultaneously. This finding underscores a critical challenge in developing versatile and efficient AI systems unveiled through the analysis of neural networks using topological representations, an approach akin to CNT that examines a network structure and connectivity.
In continuity with Petri et al., Saxe et al.~\shortcite{SAXE2022} introduce the Gated Deep Linear Network model. Through this model, they reveal that the learning process in structured networks is a `neural race' in which different components compete to learn and represent information. This process is biased towards forming structured representations of knowledge, which determine a DNN's ability to generalise, transfer knowledge and handle multiple tasks simultaneously.
La Malfa et al.~\shortcite{LAMALFA2021} study the DNNs training dynamics through the lens of CNT. Their approach is local and global in that it studies single networks and populations of evenly initialised DNNs. They formalise metrics for weights, neurons and hidden layers, yet their measures do not account for the value of the input. Scabini et al.~\shortcite{SCABINI2022} unveil the topological properties of a broad range of FCs and the benefits of random weight initialisation. They introduce Bag-Of-Neurons, a technique designed to identify topological signatures to group similar neurons (i.e., those dedicated to solving a specific sub-problem of a task such as edge detection).

\section{Methodology}~\label{S3}
First, we provide background notation to describe a DNN via CNT metrics~\cite{TESTOLIN2019,LAMALFA2021}: the reference architecture is an FC network that solves a classification task. We then introduce CNT metrics that account for the input data via sampling from the training distribution. Finally, we show how to compute these metrics for a broad range of neural network architectures, including CNNs and RNNs.

\subsection{Background}
DNNs as graphs have three main components: the \emph{input layer}, which is the `receptor' of the input data (e.g., pixels of an image, audio samples, textual data, etc.), the \emph{hidden layers}, which are stacked layers of neurons that transform the input via affine transformation followed by non-linear function activations, and the \emph{output}, e.g., the network's classification.
Formally, we consider an FC network that solves a supervised classification task, i.e., the network learns an input-output mapping $f: \mathbb{R}^d \xrightarrow{} \mathbb{R}^m$ that minimises a generic loss function $\mathcal{L}(f(x), y)$, between each input-output pair $(x,y)$. An input $x$ is a $d$-dimensional vector $x \in \mathbb{R}^d$ drawn from a distribution, while each corresponding output is either from a discrete set in case of classification, i.e., $c \in C \ . \ |C|=m$, or it is continuous in case of regression, i.e., $y \in \mathbb{R}^m$.
An FC architecture consists of $L>0$ dense layers stacked together, each of a variable number of neurons: within each hidden layer $\ell$, a neuron $n^{[\ell]}_{i}$ is connected through a weighted link to all the neurons of the successive layer $\ell + 1$.
The output $z^{[\ell]}$ of a layer $\ell$ is the product of an affine transformation between a matrix of weights $\Omega^{[\ell]}$, plus eventually a bias term $\beta^{[\ell]}$, namely $z^{[\ell]} \ = \ z^{[\ell-1]}\Omega^{[\ell]} \ + \ \beta^{[\ell]}$, followed by a non-linear activation function $f^{[\ell]}(z^{[\ell]})$. For an FC network, $\Omega^{[\ell]}$ is a matrix of size $\mathcal{N}^{[\ell]}\times\mathcal{N}^{[\ell+1]}$ and $\beta^{[\ell]}$ is a vector of size $\mathcal{N}^{[\ell+1]}$.
We denote the input and output vectors as $x=z^{[0]}$ and $y = z^{[L]}$, while $\Omega^{[\ell]}$ and $\beta^{[\ell]}$ refer to the parameters of a neural network layer $\ell$. The output of the neural network is defined as $y=f^{[L]}(z^{[L]}) = z^{[L-1]}\Omega^{[L]} \ + \ \beta^{[L]}$. For a classification task, the output is a vector of real numbers $y \in \mathbb{R}^m$, from which the $argmax$ operator extracts the predicted class. We refer to the input-output relation of a neural network at layer $\ell$ as $z^{[\ell]} = \mathbf{f}(x, \ \Omega^{[:\ell]}, \ \beta^{[:\ell]})$.

\subsection{Complex Networks Metrics for DNNs}
A DNN can be represented as a set of nodes (neurons) connected by weighted edges. This intuition is sufficient to describe a DNNs via CNT.
Formally, a DNN is a directed bipartite graph $G = \langle N, E\rangle$, where each node $n^{[\ell]}_{i} \ \in \ N$ corresponds to a neuron in the $\ell$-th hidden layer. The intensity of a connection is a real number $\omega^{[\ell]}_{i,j}$ assigned to an edge $(e_{n^{[\ell]}_{i}, n^{[\ell+1]}_{j}} \ \in \ E)$ that connects two neurons. 

\paragraph{Link Weights.}
The Link Weights provide insight into how weights and biases adapt during training. Standard measurements of such metrics are the weights mean and variance at each layer during training. For DNN at layer $\ell$ they are defined as:
\begin{equation}
\mu^{[\ell]} = \dfrac{1}{N^{[\ell]} N^{[\ell+1]} }\sum_{i=1}^{N^{[\ell]}} 
\sum_{j=1}^{N^{[\ell+1]}} \omega^{[\ell]}_{i,j} + \beta^{[\ell]}_i
\end{equation}
\begin{equation}
\delta^{[\ell]} = \dfrac{1}{N^{[\ell]} N^{[\ell+1]} }\sum_{i=1}^{N^{[\ell]}} 
\sum_{j=1}^{N^{[\ell+1]}} ((\omega^{[\ell]}_{i,j} + \beta^{[\ell]}_i ) -\mu^{[\ell]})^2
\end{equation}
Monitoring weights mean and variance throughout training provides insights into the learning process's effectiveness and stability. If the norm of the weights does not increase, it could indicate over-regularisation in the model. Conversely, excessively large weight values might lead to overfitting.

\paragraph{Nodes Strength.}
Formally, the strength $s^{[\ell]}_{k}$ of a neuron $n^{[\ell]}_{k}$ is the sum of the weights of the edges incident in $n^{[\ell]}_{k}$. Since neural network graphs are directed, two components contribute to the Node Strength: the sum of the weights of outgoing edges $s^{[\ell]}_{out,k}$, and the sum of the weights of in-going links $s^{[\ell]}_{in,k}$.
\begin{equation} \label{eq:nodes-strength}
s^{[\ell]}_{k} = s^{[\ell]}_{in,k} + s^{[\ell]}_{out,k} =\ \sum_{i=1}^{N^{[\ell]}}(\omega^{[\ell]}_{i,k} + \beta^{[\ell]}_k) + \sum_{j=1}^{N^{[\ell+1]}}\omega^{[\ell+1]}_{k,j}
\end{equation}
Node strength reflects how strong a connection is for a specific feature. 
In literature, the Node Strength account for both in- and out-coming edges~\cite{LAMALFA2021}; we will consider them separately.

\paragraph{Layers Fluctuation.}
\emph{Layers Fluctuation} extends the idea of Nodes Fluctuation~\cite{porta2006network} to measure the variability of metrics at the level of a network's hidden layers.
CNT classically defines Nodes Disparity as $Y^{[\ell]} = \sum_{i=1}^{N^{[\ell]}}[{\omega^{[\ell]}_{i}}/{s^{[\ell]}_{i}}]^2$. However, weights generally have positive and negative values that could cancel each other out. 
To address this issue, we propose a metric designed for DNNs that captures the strength fluctuations within each layer, reflecting the interactions among nodes at the same depth.
The Layers Fluctuation is defined as:
\begin{equation}
Y^{[\ell]} = \sqrt{\dfrac{\sum_{i=1}^{N^{[\ell]}}(s^{[\ell]}_{i} - \hat{s}^{[\ell]})^2}{I}}
\end{equation}
where $\hat{s}^{[\ell]}$ is computed as the average value of Nodes Strength at layer $\ell$, namely $\hat{s}^{[\ell]}=\dfrac{1}{N^{[\ell]}}\sum^{m}_{i=1}s^{[\ell]}_{i}$, being $N^{[\ell]}$ the number of nodes/neurons at layer $\ell$. 
This metric identifies asymmetries and disparities in the network at the layer level rather than focusing on individual nodes.
Unlike the standard Nodes Fluctuation, Layers Fluctuation characterises the dynamics of an entire layer and accounts for bottlenecks within the network architecture.

\subsection{Data-dependent CNT Metrics}
We now introduce and formalise CNT metrics that account for the value of the input data, namely the \emph{Neurons Strength} and \emph{Neurons Activation}. 
Specifically, the training data distribution informs the computation of \emph{Neurons Strength} and \emph{Neurons Activation} so that for two different datasets, the results would vary accordingly (while it is not the case for the previous metrics).
We then discuss how to compute all the above CNT metrics, including \emph{Nodes Strength} and \emph{Layer Fluctuation} to CNNs and RNNs. 

\paragraph{Neurons Strength.}
The mathematical formulation of Neurons Strength for a neuron $n^{[\ell]}_{k}$ in layer $\ell$ is given by:
\begin{multline}
\zeta^{[\ell]}_{k} =\ \sum_{i=1}^{N^{[\ell]}}z^{[\ell-1]}_{i}\omega^{[\ell]}_{i,k} + \beta^{[\ell]}_{k}, \\ z^{[\ell-1]} = \mathbf{f}(x, \Omega^{[:\ell]}, \beta^{[:\ell]}) \ . \ x \sim \mathcal{X}
\end{multline}
Here, $\zeta^{[\ell]}_{k}$ represents the strength of neuron $k$ in layer $\ell$, considering the effects of both the activation functions of previous layers and the input values drawn from a distribution $\mathcal{X}$. This approach provides a more comprehensive understanding of the neuron's role and influence within the network, factoring in the data being processed.

\paragraph{Neurons Activation.}
In Deep Neural Networks (DNNs), each neuron's activation level is determined by both the input values and the specific activation functions used in the network. The activation of a neuron in layer $\ell$ can be mathematically expressed as:
\begin{multline}
a^{[\ell]}_{k} =\ f^{[\ell]}(\sum_{i=1}^{N^{[\ell]}}z^{[\ell-1]}_{i}\omega^{[\ell]}_{i,k} + \beta^{[\ell]}_{k}), \\ \ \ z^{[\ell-1]} = \mathbf{f}(x, \Omega^{[:\ell]}, \beta^{[:\ell]}) \ . \ x \sim \mathcal{X}.
\end{multline}
This equation highlights how the activation value $a^{[\ell]}_{k}$ of a neuron depends on the weighted sum of activations from the previous layer, adjusted by the neuron's weights and bias, and then transformed by the activation function $f^{[\ell]}$.
A neuron exhibiting an unusually high Node Strength transmits a stronger \emph{signal} than others. In such a scenario, an input does not elicit all the neurons uniformly, i.e., a neuron conveys more significant information for the classification task. Conversely, a neuron that transmits a weak \emph{signal} might be a candidate for pruning, which can help reduce the overall complexity of the network without significantly impacting the layer's output.

\subsection{CNT Metrics for Convolution and Recursion}\label{sec:cnn-metrics}
\begin{figure}
    \centering
    \includegraphics[width=1\linewidth]{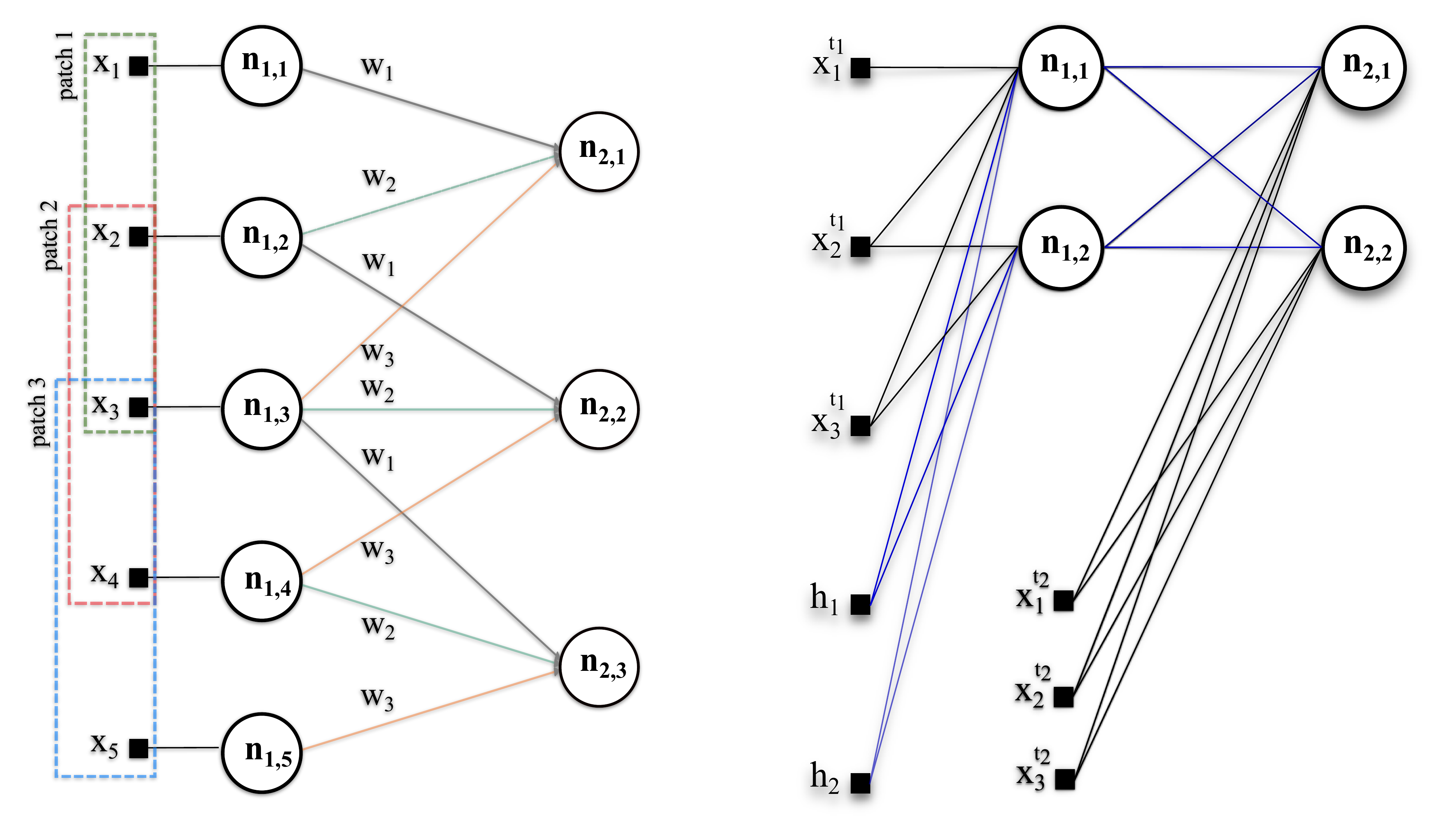}
    \caption{For CNNs, CNT metrics are computed by isolating each input patch and the kernel responsible for a dot-product in a layer (left), while for RNNs, metrics can be computed by unfolding each input feature through time (right).}
    \label{cnt-cnn-rnn}
\end{figure}
The role of architectural inductive biases in enhancing the performance of deep learning models is well-established and supported by decades of extensive research. These biases, integrated into a model architecture, have significantly influenced the field of artificial intelligence. For instance, CNNs were designed with biases towards local connectivity, inspired by the human visual system~\cite{LECUN1995}. Similarly, the development of recurrent networks, particularly the introduction of gates and memory cells in Long Short-Term Memory (LSTM) networks~\cite{hochreiter1997long}, exemplifies how these biases enable the retention and accessibility of information over extended time steps. 
In this section, we adapt CNT metrics to CNN and RNN architectures, thus moving beyond the model of reference in most works on CNT applied to DNNs.

\paragraph{Convolutional Neural Networks.} 
Convolution is the building block operator in CNNs. An input matrix of size $w \times h$ (e..g, an image represented as a grid of pixels) is split into (possibly overlapping) patches of size $k \times k$. A point-wise multiplication between each patch and the kernel, which has the same size as each patch, shrinks the input and produces an output that is then activated and fed to the next DNN's layer.
In mathematical terms, for an input matrix $z_{\ell}$ of dimensions $w \times h$ and a kernel $\Omega$ of size $m \times m$ (with $m \le w$ and $m \le h$), the convolution operation can be represented as $z'= z * \Omega$.
We now report the formula to compute the Neurons Strength for a CNN layer to highlight the differences with the Fully Connected case. For the Neurons Activation, the operations are analogous.
\begin{multline}
\zeta^{[\ell]}_{k} =\ z^{[\ell-1]}_{k} * \Omega^{[\ell]} + \beta^{[\ell]}_{k}, \\ z^{[\ell-1]} = \mathbf{f}(x, \Omega^{[:\ell]}, \beta^{[:\ell]}) \ . \ x \sim \mathcal{X}.
\end{multline}
In the previous equation, $z^{[\ell-1]}_{k} * \Omega^{[\ell]}$ represents the convolution between the input patch $z^{[\ell-1]}_{k}$ and the entire kernel of weights at layer $\ell$.
One straightforward method to apply CNT metrics to this convolution operation is to transform the convolution into a dot product operation using a Toeplitz matrix. However, while this method is conceptually simple, it significantly increases both the time and space complexity from linear to quadratic, and it is thus expensive for large networks.
Our approach, therefore, is to tackle this challenge efficiently. We isolate each input portion that is element-wise multiplied by the kernel. By linking each input neuron to its corresponding output neuron (as illustrated in Figure \ref{cnt-cnn-rnn}, left), we can efficiently compute the CNT metrics for any layer. This method allows to apply complex network analysis to convolution operations without the prohibitive computational cost of the Toeplitz matrix approach.

\paragraph{Recurrent Neural Networks.} 
RNNs are designed to process sequential input, where the output of the previous step influences the output at each step in the sequence. Recursion makes RNNs ideal for tasks involving temporal data, like speech recognition or language modelling.
The output of a single-layer recurrent RNN for the $t>0$ temporal dimension of $x$ is the following:
\begin{equation}
h^{(t+1)} =\ \mathbf{f}(z^{(t)}\Omega + h^{(t)}U).
\end{equation}
In the previous equation, where the bias term is omitted for clarity, $U$ represents a matrix of trainable parameters used to process the hidden unit $h^{(t)}$. Usually the value of $h^{(1)}$ is initialised to zero.
To adapt CNT metrics for RNNs, we `unfold' each input feature along its temporal dimension and treat each recurrent step as a layer-wise multiplication in an FC topology.
This transformation is computationally less efficient than relying on symbolic loops, yet it is necessary to compute the CNT metrics for each input feature.
The closed form formula to compute the Neurons Activation for an RNN Cell with one hidden layer, at time $t>0$, and with parameters $\Omega$ and $U$ (as sketched in Figure~\ref{cnt-cnn-rnn}, right) is the following:
\begin{multline}
\zeta^{(t)} =\ x^{(t-1)}\Omega + \mathbf{f}(\dots \mathbf{f}(x^{(1)}\Omega + h^{(1)}U) \dots) U  \\ . \ x \sim \mathcal{X}.
\end{multline}
In the previous equation, the value of the Neuron Strength is computed by first sampling an input $x$ from the input data distribution $\mathcal{X}$, then by recursively unrolling the RNN and processing each temporal feature $x^{(j)}$ to compute the value of the hidden unit $h^{(j)}$ at times $j < t$.
In the experiments, we show how the Neuron Strength can enhance a model's interpretability of which input features elicit the most neurons in an RNN recursive layer.

\section{Experiments}~\label{S4}
\begin{figure*}[!ht]
    \centering
    \includegraphics[width=0.9\linewidth]{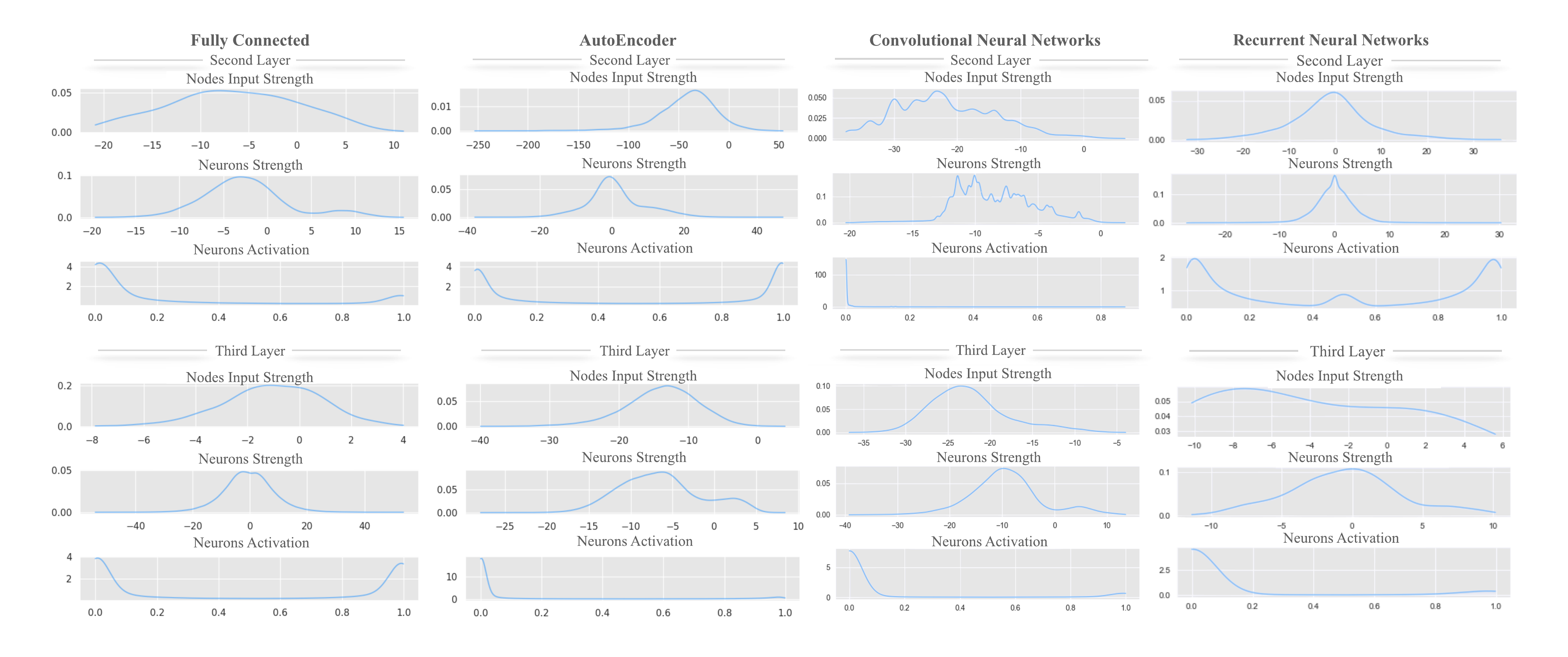}
    \caption{Analysis of CNT metrics across the second and third layers (hidden and output layers) for three-layer depth FCs, CNNs, RNNs and AEs on the MNIST dataset. Each column corresponds to an architecture, and the figures illustrate the distribution functions computed on a pool of $30$ neural networks trained on the task.}
    \label{exp:architectures}
\end{figure*}
\begin{figure*}[!ht]
    \centering
    \includegraphics[width=0.9\linewidth]{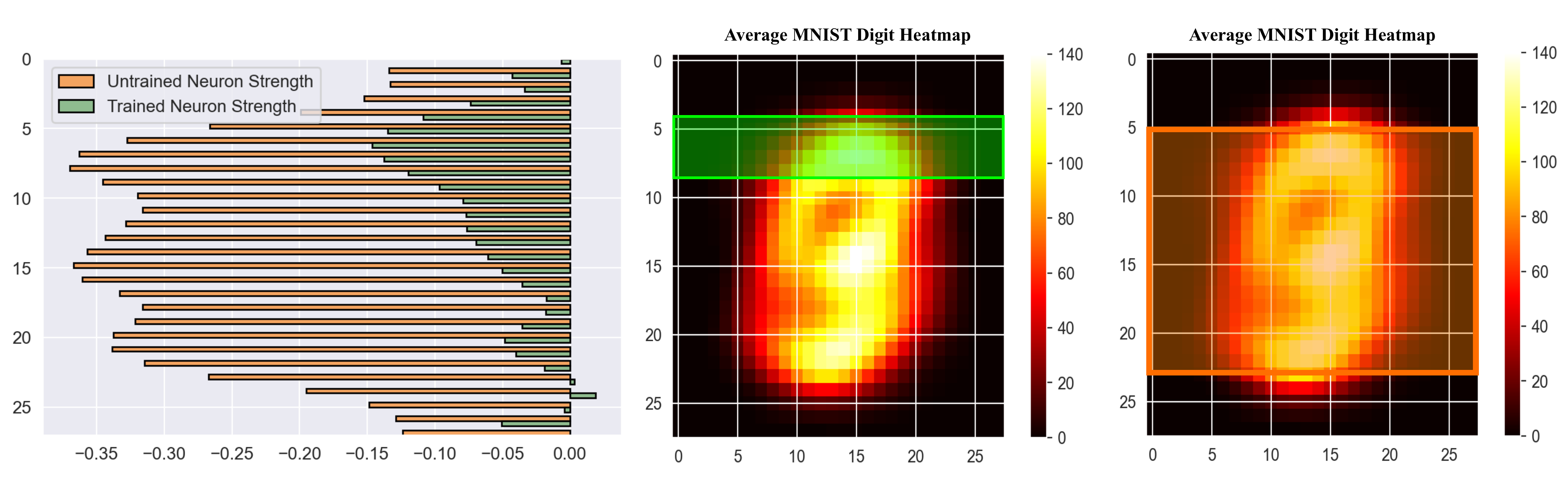}
    \includegraphics[width=0.9\linewidth]{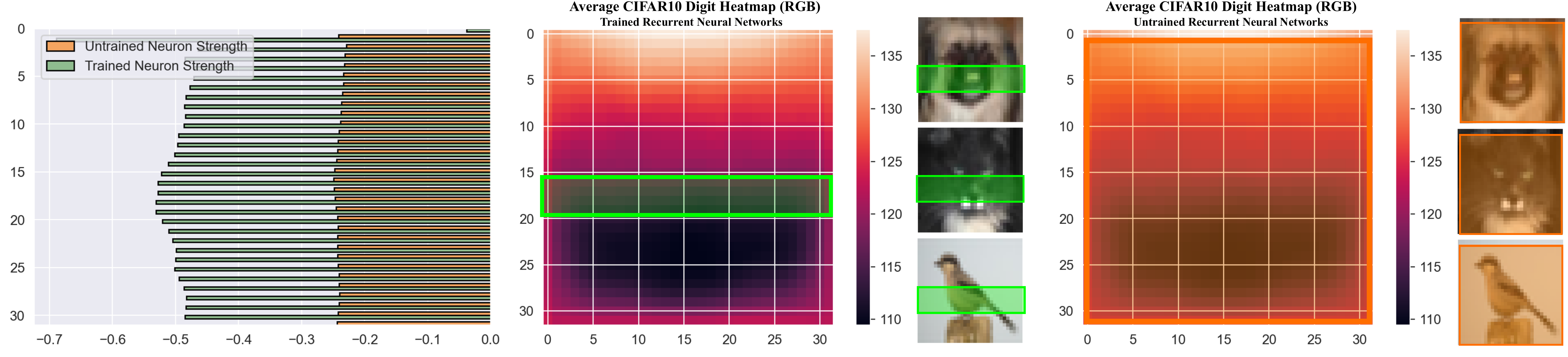}
    \caption{Neurons Strength metric for $30$ RNNs on MNIST (top) and CIFAR10 (bottom) classification tasks, distinguishing between networks that have been \textcolor{green}{trained} and those that remain \textcolor{orange}{untrained}. The left side of the figure quantifies the Neurons Strength, while the right side visualises a global heatmap of neurons that are most elicited by MNIST/CIFAR10 inputs.}  
    \label{exp:rnn-mnist}
\end{figure*}
\begin{figure*}[!ht]
    \centering
    \includegraphics[width=0.9\linewidth]{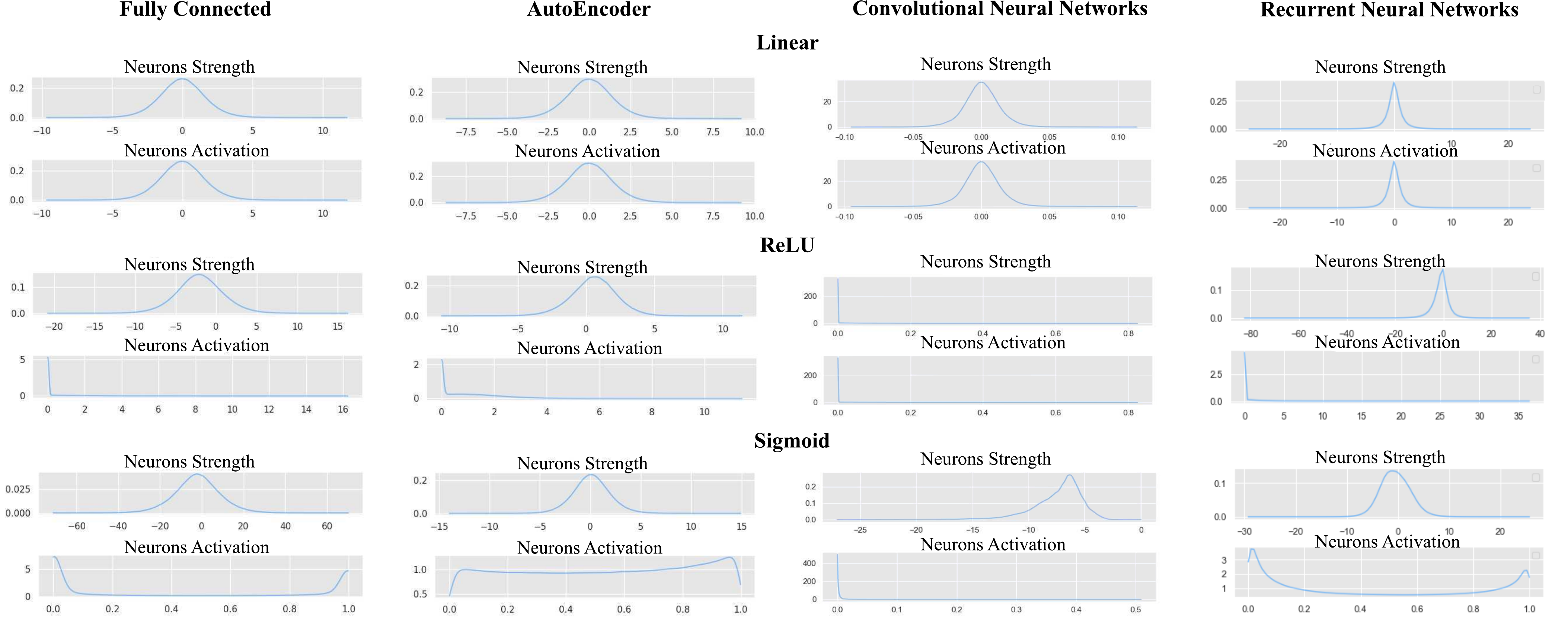}
    \caption{Analysis of CNT metrics for three-layer depth FCs, CNNs, RNNs and AEs on the CIFAR10 dataset and different activation functions (linear, ReLU and sigmoid). Each column corresponds to an architecture, and the figures illustrate the distribution functions computed on a pool of $30$ neural networks trained on the task.}
    \label{exp:activations}
\end{figure*}
In this section, we conduct experiments to assess to which extent CNT identifies patterns in DNNs: we define three complementary levels of analysis.
The first level (I) aims to distinguish dominating CNT patterns for architecturally similar networks: we train on MNIST and CIFAR10 three-layer depth FCs, CNNs, RNNs and AEs equipped with the same activation functions and a comparable number of parameters.\footnote{All the details on the accuracy of the networks are reported in the Appendix.} While we test FCs, CNNs, and RNNs on image classification, AEs are trained to compress and reconstruct the input.
As we keep the architectures as simple as possible, MNIST results in a relatively simple task where all the models perform well, while on CIFAR10, CNNs have better performances aided by their inductive bias towards image classification.
The second level (II) studies the incidence of different activation functions (i.e., linear, ReLU, sigmoid) on CNT metrics for FCs, CNNs, RNNs and AEs. Similarly to the previous setting, the networks are architecturally similar in terms of hidden layers and number of trainable parameters, yet differ in their activation functions.
The third level of analysis (III) explores the impact of depth (i.e., the number of hidden layers) on CNT metrics. Deeper neural networks learn more complex features from data and positively correlate with a DNN's performance on the task. Also, deeper layers can learn increasingly abstract representations of the data. In image processing, initial layers might detect edges and textures, while deeper layers might identify more complex patterns or objects. We investigate and report the impact of depth by varying the number of hidden layers of FCs and AEs from three to seven (while further analyses are reported in the code repository).

We conduct all the experiments on two standard datasets in pattern recognition and computer vision, namely MNIST and CIFAR10~\cite{LECUN1995,krizhevsky2010cifar}. While relatively simple, these two benchmarks set a baseline for future analyses on more complex tasks; we stress that our framework can be applied to any dataset a DNN can tackle.
Generally, we compute all the CNT metrics on a $30$ trained neural network pool. We initialise the weights of each DNN via sampling from a Gaussian distribution of known variance between $0.05$ (MNIST) and $0.5$ (CIFAR10).\footnote{Higher variance in the parameter initialisation can sometimes help the network to escape poor local minima early in training, which are more prevalent in datasets like CIFAR10.}

\paragraph{I. CNT metrics of different architectures.}
\begin{figure*}[!ht]
    \centering
    \includegraphics[width=0.8\linewidth]{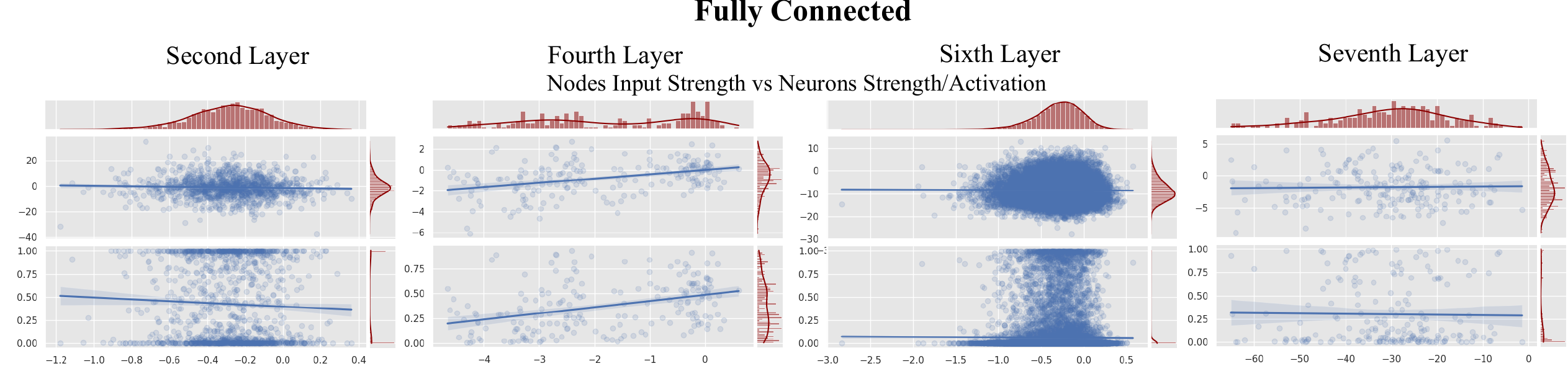}
    \includegraphics[width=0.8\linewidth]{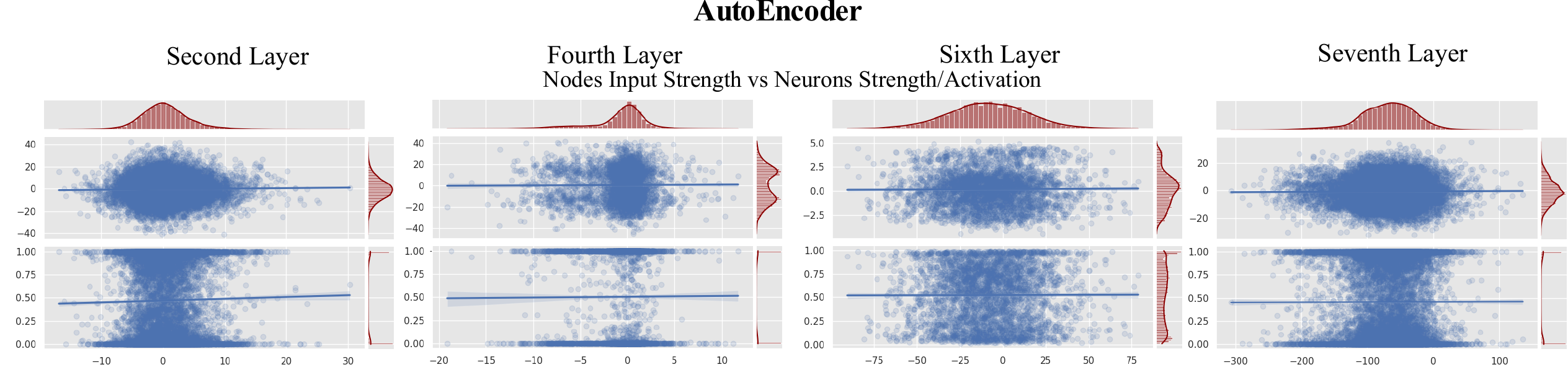}
    \caption{Neurons Strength and Activation, and scatter-plot of the correlation between Nodes Strength and Neurons Strength and Activation, for seven-layer depth FCs and AEs. The figures illustrate the distribution functions computed on a pool of $30$ neural networks trained on the task.}
    \label{exp:depth-7l}
\end{figure*}
MNIST images consist of white-scaled digits on a dark background. As reported in the literature, FC networks tend to present `dead units', i.e. an abundance of negative or close to zero weights, a hint of an over-parametrised network~\cite{TESTOLIN2019}. Our results confirm this trend for FCs and extend it to all the other architectures (CNNs, RNNs and AEs), as we show in Figure~\ref{exp:architectures}; yet, for AEs, the phenomenon is more remarked, with a probability distribution of Neurons Strength and Activation peaked respectively around negative values and at the extremes of the distribution. We further notice that AEs compress information by design, i.e., a `bottleneck', as shown in the third layer of Figure~\ref{exp:architectures}. 
On the other hand, CNNs' Nodes Strength and Neurons Strength are multimodal, with spikes corresponding to the patterns a network learns to perform local edge detection. We hypothesise that CNNs specialise neurons within the same layer to activate them for different input values. For example, the spikes may represent the information relative to the number of edges in a digit, as that is one of the sub-tasks that requires more information to be encoded.
Globally, the influence of positive and negative weights in FC, AEs and RNNs seems balanced, while CNNs exhibit a profusion of negative weights. This suggests that, during the training phase, CNNs may prioritise different input features than FCs, RNNs and AEs.
RNNs' Neuron Strength distribution exhibits a pronounced Kurtosis, a feature that isn't observed when analysing Node Input Strength alone and is thus imputable to the data distribution. Additionally, the Neuron Activation displays a unique pattern: unlike other architectures, the density is distributed not just at the extremes of the sigmoid function, but also significantly in the middle (between 0.4 and 0.6), a sign that recurrent architectures are less likely to settle at the extremes, resulting in more varied activation patterns. For reasons of space, results for CIFAR10 are reported in the Appendix and the code repository.
Regarding interpretability, we further investigate the dynamics of the Neurons Strength in an RNNs' autoregressive loop.
This approach involves an RNN sequentially processing a segment of the input, updating its hidden states, and ultimately classifying the image. 
In Figures \ref{exp:rnn-mnist}, we compare the Neurons Strength of trained and untrained RNNs on the MNIST and CIFAR10 datasets. In both cases, the Neurons Strength of trained RNNs localises a specific image region that elicits the neurons the most, while that doesn't occur for untrained networks. This suggests that the training phase calibrates a model to activate mostly on specific patterns located in the central part of the input.

\paragraph{II. Incidence of the activation function.}
Non-linear activation functions allow models to learn complex patterns in data. We compare the effects of linear, ReLU, and sigmoid activation functions for FCs, CNNs, RNNs and AEs on CIFAR10. 
Linear activations, lacking the ability to model non-convex optimisation landscapes, typically lead to Neurons Strength and Neurons Activation distributions centred around zero. In contrast, ReLU and sigmoid activations result in asymmetric distributions, favouring negative values, often correlating with enhanced accuracy. Notably, ReLU's inherent asymmetry results in distributions skewed more significantly towards extreme negative values than those observed with sigmoid, as we report in Figure~\ref{exp:activations} (first column).
AEs with linear and ReLU activations show similar behaviours to FCs. However, models equipped with sigmoid exhibit Neuron Activation distributed evenly around its support. This phenomenon suggests the AEs correctly reconstruct the input data with dynamics that diverge from FCs, CNNs and RNNs, which solve a different task (classification).
For linear models, CNNs' Neurons Activation centres around zero, with low variance compared to other architectures. Conversely, non-linear activations skew the distribution of the metrics toward zero. Both ReLU- and sigmoid-activated networks leverage negatively activated neurons to discriminate between different classes. This phenomenon is remarkably different from what happens with FCs and RNNs where the Neurons Activation accumulates around the extreme values of the activation (zero and one), suggesting that the decision rules encoded internally by CNNs may diverge from that of other architectures.  

\subsection{III. Incidence of neural network depth.}
\begin{figure}[!ht]
    \centering
    \includegraphics[width=1\linewidth]{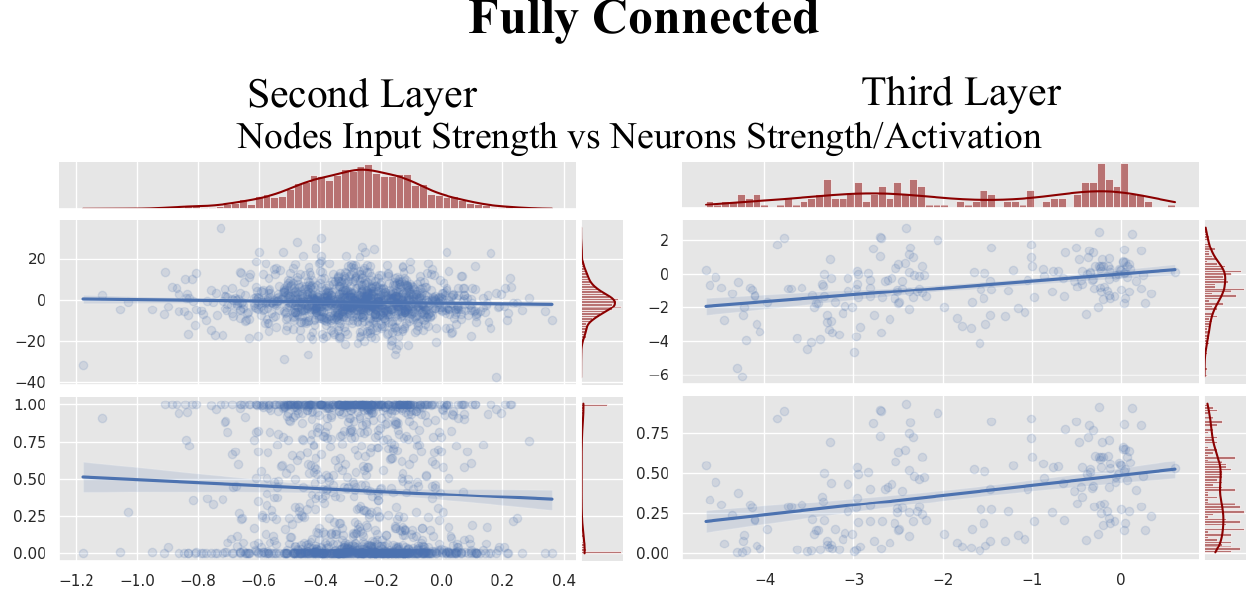}
    \includegraphics[width=1\linewidth]{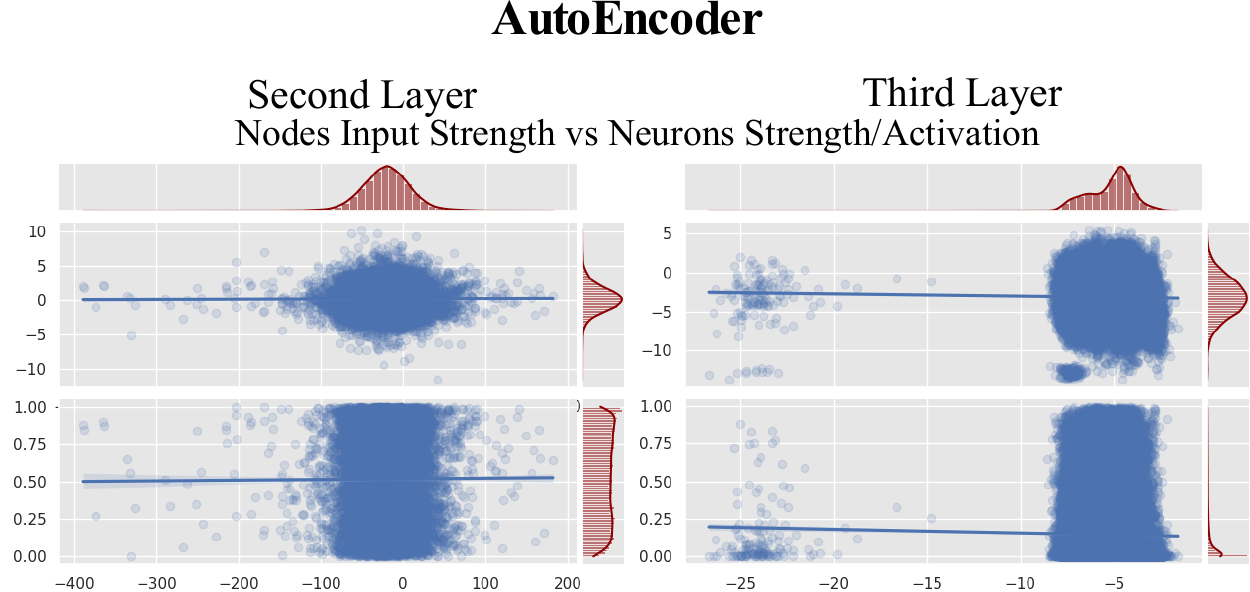}
    \caption{Neurons Strength and Activation, and scatter-plot of the correlation between Nodes Strength and Neurons Strength and Activation, for three-layer depth FCs and AEs. The figures illustrate the distribution functions computed on a pool of $30$ neural networks trained on the task.}
    \label{exp:depth-3l}
\end{figure}
We conclude with an analysis of the effect of depth on the dynamics of CNT metrics on CIFAR10. We compare FCs and AEs with three and seven layers, respectively. While FCs perform classification, AEs are trained to reconstruct the input with minimal data loss.\footnote{Results for the other architectures on MNIST and CIFAR10, for five and nine layers, are reported in the code repository.} 
In Figure~\ref{exp:depth-3l}, we report the Neurons Strength and Activation, which we further correlate with the Nodes Strength. Shallow FC architectures operate with similar dynamics as deeper networks, with the second and third layers of Figure~\ref{exp:depth-3l} (top) that overlap with respectively the fourth and seventh of Figure~\ref{exp:depth-7l} (top): in conclusion, deeper networks exhibit more complex patterns in their hidden layers, that shallow lack. For this task, deeper networks leverage the same `building block' of shallow networks to build increasingly complex input representations and solve the task more accurately. 
Conversely, the dynamics of shallow and deep AEs diverge sensibly: seven-layer AEs seem not to leverage existing `building blocks' of shallow architectures, suggesting that data compression is a dynamic process heavily influenced by the number of layers in a network. 
We also notice that the Nodes Strength does not correlate with the Neurons Strength or Activation, supporting the development of CNT metrics that incorporate the effect of data in their dynamics.

\section{Conclusions and Future Works}
This paper introduces a unified framework for representing neural networks via CNT, incorporating new metrics influenced by input data and enhancing traditional topological CNT analysis. Our extensive experiments on FCs, CNNs, RNNs, and AEs reveal distinct dynamics and training patterns across architectures, activation functions, and depths. our key findings include over-parametrisation in most architectures, especially with the MNIST task, and CNNs demonstrating localised learning through multimodal Nodes and Neurons Strength. Non-linear activations in models lead to asymmetric distributions, causing complex pattern learning. AEs' Neurons Activation distribution in deeper models suggests their learning dynamics do not leverage the same dynamics as shallow architectures. Unlike FCs and RNNs, CNNs with non-linear activations learn discriminative patterns for classification, mostly in negative regions of the activation function.
In future works, we will extend this framework to advanced architectures, both theoretically and empirically. In particular, we believe that CNT can contribute to interpreting self-attention~\cite{vaswani2017attention}, an architecture that powers language and computer vision models.
This work is also meant to encourage other researchers with expertise in both machine learning and physics to contribute to the formalisation of learning systems as graphs to unveil their training dynamics as the current state-of-the-art approaches do not study the evolution of untrained networks during the learning phase. 

\section*{Acknowledgments}
G.L.M.'s work was supported by UK Research and Innovation [grant number EP/S023356/1], in the UKRI Centre for Doctoral Training in Safe and Trusted Artificial Intelligence (\url{www.safeandtrustedai.org}). 

G.N. thanks the University of Cambridge for the extended research periods and the opportunity to be a Visiting Professor during the Sabbatical year. Moreover, GN thanks Wolfson College, the Department of Biochemistry, and the Cambridge Systems Biology Centre for the hospitality and warm welcome that allowed for peaceful studies, novel, interesting collaborations, and fruitful research activities.

V.L. acknowledges support from the European Union, NextGenerationEU, GRINS project (GRINS PE00000018 - CUP E63C22002120006). 

\clearpage
\bibliographystyle{named}
\bibliography{ijcai24}

\clearpage

\appendix
\section{Suppementary Results}
\subsection{Accuracy of Trained Neural Networks}
In Table~\ref{tab:accuracy}, we report the training details of FCs, CNNs, RNNs and AEs architectures.
\begin{table*}[!ht]
\centering
\begin{tabular}{|c|c|c|c|c|c|l|}
\hline
\multicolumn{1}{|l|}{} &
  \textbf{Number of Networks} &
  \textbf{Sample Size} &
  \textbf{Layers} &
  \textbf{\begin{tabular}[c]{@{}c@{}}Performance\\ MNIST\end{tabular}} &
  \textbf{\begin{tabular}[c]{@{}c@{}}Performance\\ CIFAR10\end{tabular}} &
  \multicolumn{1}{c|}{\textbf{Reference}} \\ \hline
\textbf{FC}  & 30 & 100 & 3, 5, 7, 9 & 0.90 - 0.95           & 0.27 - 0.41 & Figures~\ref{exp:architectures},~\ref{exp:activations},~\ref{exp:depth-7l},~\ref{exp:depth-3l},~\ref{exp:fc-ae-9l}   \\ \hline
\textbf{CNN} & 30 & 100 & 3, 5       & 0.89 - 0.96           & 0.54 - 0.60 & Figures~\ref{exp:architectures},~\ref{exp:activations} \\ \hline
\textbf{RNN} & 30 & 100 & -          & 0.84 - 0.89           & 0.43 - 0.45 &  Figures~\ref{exp:architectures},~\ref{exp:rnn-mnist},~\ref{exp:activations} \\ \hline
\textbf{AE}  & 30 & 100 & 3, 5, 7, 9 & 0.058 - 0.067         & 0.83 - 0.93       &  Figures~\ref{exp:architectures},~\ref{exp:activations},~\ref{exp:depth-7l},~\ref{exp:depth-3l},~\ref{exp:fc-ae-9l}  \\ \hline
\end{tabular}
\caption{Architectural details and performance of different architectures on MNIST and CIFAR10 datasets. For FCs, CNNs and RNNs, we report the overall minimum and maximum accuracy on $30$ independent experiments (independently from the architecture and activation). For the AEs, we report the value of the MSE reconstruction loss (the lower, the better). The sample size refers to the number of samples per network used to compute Neurons Strength and Activations.}
\label{tab:accuracy}
\end{table*}

\subsection{Additional Results}
\begin{figure*}[!ht]
    \centering
    \includegraphics[width=0.8\linewidth]{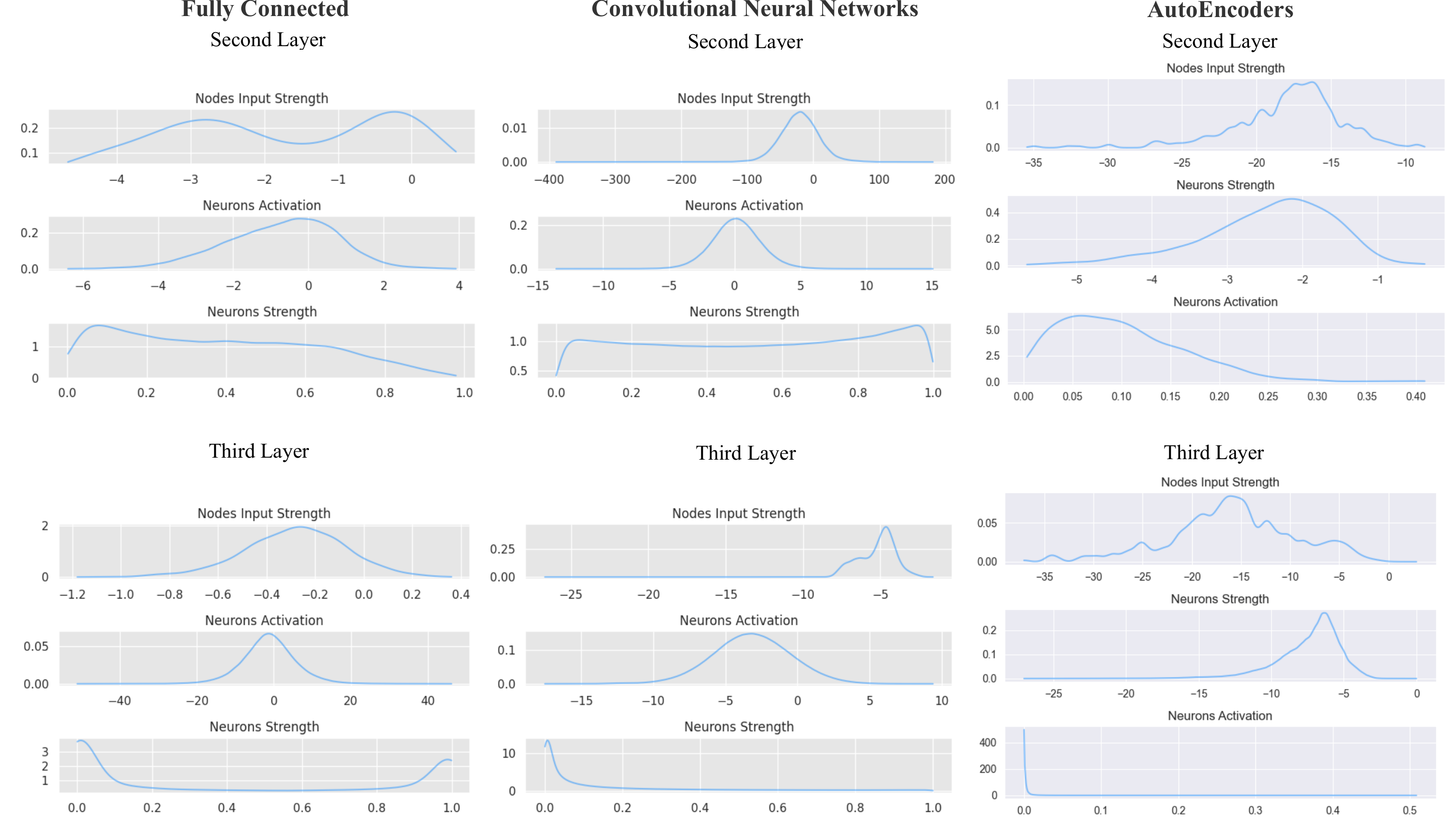}
    \caption{Neurons Strength and Activation, and scatter-plot of the correlation between Nodes Strength and Neurons Strength and Activation, for three-layer depth FCs, CNNs and AEs on the CIFAR10 dataset. The figures illustrate the distribution functions computed on a pool of $30$ neural networks trained on the task.}
    \label{exp:fc-cnn-ae-3l-cifar}
\end{figure*}

\begin{figure*}
    \centering
    \includegraphics[width=0.9\linewidth]{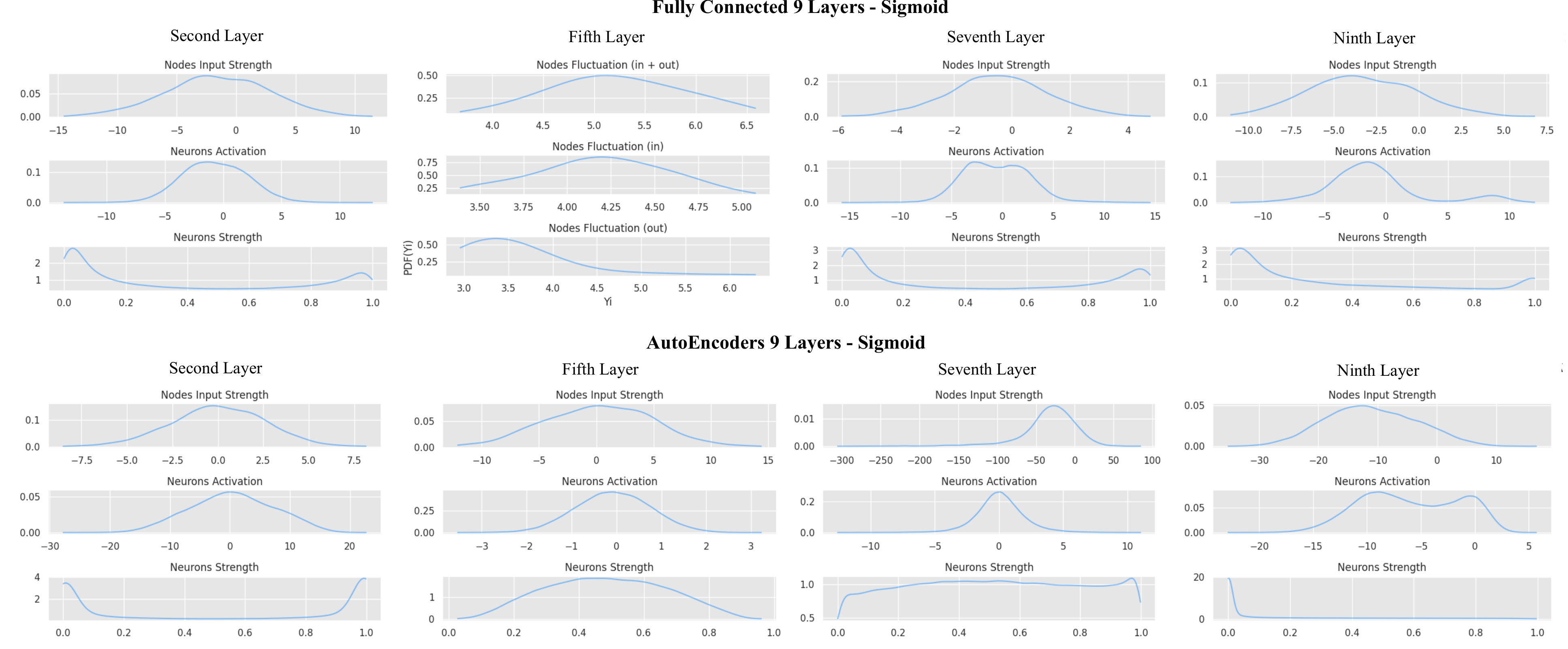}
    \caption{Analysis of CNT metrics across the second, fifth, seventh and ninth layers for nine-layer depth FCs and AEs on the MNIST dataset. Each column corresponds to an architecture, and the figures illustrate the distribution functions computed on a pool of $30$ neural networks trained on the task.}
    \label{exp:fc-ae-9l}
\end{figure*}

In this section, we report some results that complement the main paper.
In the code repository, we further include many experiments on different architectures, depths and datasets that, for reason of space, do not fit with the main scope of the article.
In particular, Figure~\ref{exp:fc-cnn-ae-3l-cifar} shows different activation patterns for three-layer FCs, CNNs and AEs on the CIFAR10 dataset, while Figure~\ref{exp:fc-ae-9l} reports different activation patterns for nine-layer FCs and AEs on the MNIST dataset. 

\end{document}